\documentclass[10pt,twocolumn,letterpaper]{article}
\usepackage{booktabs}
\usepackage[table]{xcolor}
\usepackage{tabularx}
\usepackage{adjustbox}
\usepackage{booktabs}   
\usepackage{multirow}

\usepackage[pagenumbers]{cvpr} 

\definecolor{cvprblue}{rgb}{0.21,0.49,0.74}
\usepackage[pagebackref,breaklinks,colorlinks,allcolors=cvprblue]{hyperref}

\def\paperID{6627} 
\def\confName{CVPR}
\def\confYear{2026}

\def\algorithmname{MoE3D}

\title{MoE3D: Mixture of Experts meets Multi-Modal 3D Understanding}

\author{
Yu Li\textsuperscript{1} \quad
Yuenan Hou\textsuperscript{2,*} \quad
Yingmei Wei\textsuperscript{1,*} \quad
Xinge Zhu\textsuperscript{3} \quad
Yuexin Ma\textsuperscript{4} \quad
Wenqi Shao\textsuperscript{2} \quad
Yanming Guo \textsuperscript{1} \\
\textsuperscript{1}National University of Defense Technology\\
\textsuperscript{2}Shanghai AI Laboratory \\
\textsuperscript{3}The Chinese University of Hong Kong
\textsuperscript{4}ShanghaiTech University \\
\textsuperscript{*}Corresponding author
}

\begin{document}
\maketitle
\begin{abstract}
Multi-modal 3D understanding is a fundamental task in computer vision. Previous multi-modal fusion methods typically employ a single, dense fusion network, struggling to handle the significant heterogeneity and complexity across modalities, leading to suboptimal performance. In this paper, we propose \algorithmname~, which integrates Mixture of Experts (MoE) into the multi-modal learning framework. The core is that we deploy a set of specialized "expert" networks, each adept at processing a specific modality or a mode of cross-modal interaction. Specifically, the MoE-based transformer is designed to better utilize the complementary information hidden in the visual features. Information aggregation module is put forward to further enhance the fusion performance. Top-1 gating is employed to make one expert process features within expert groups, ensuring high efficiency. We further propose a progressive pre-training strategy to better leverage the semantic and 2D prior, thus equipping the network with good initialization. Our \algorithmname~achieves competitive performance across four prevalent 3D understanding tasks. Notably, our \algorithmname~surpasses the top-performing counterpart by \textbf{6.1} mIoU on Multi3DRefer. Codes will be available upon publication.
\end{abstract}    
\section{Introduction}
\label{sec:intro}

\begin{figure*}[!t]
    \centering
    \includegraphics[width=0.95\linewidth]{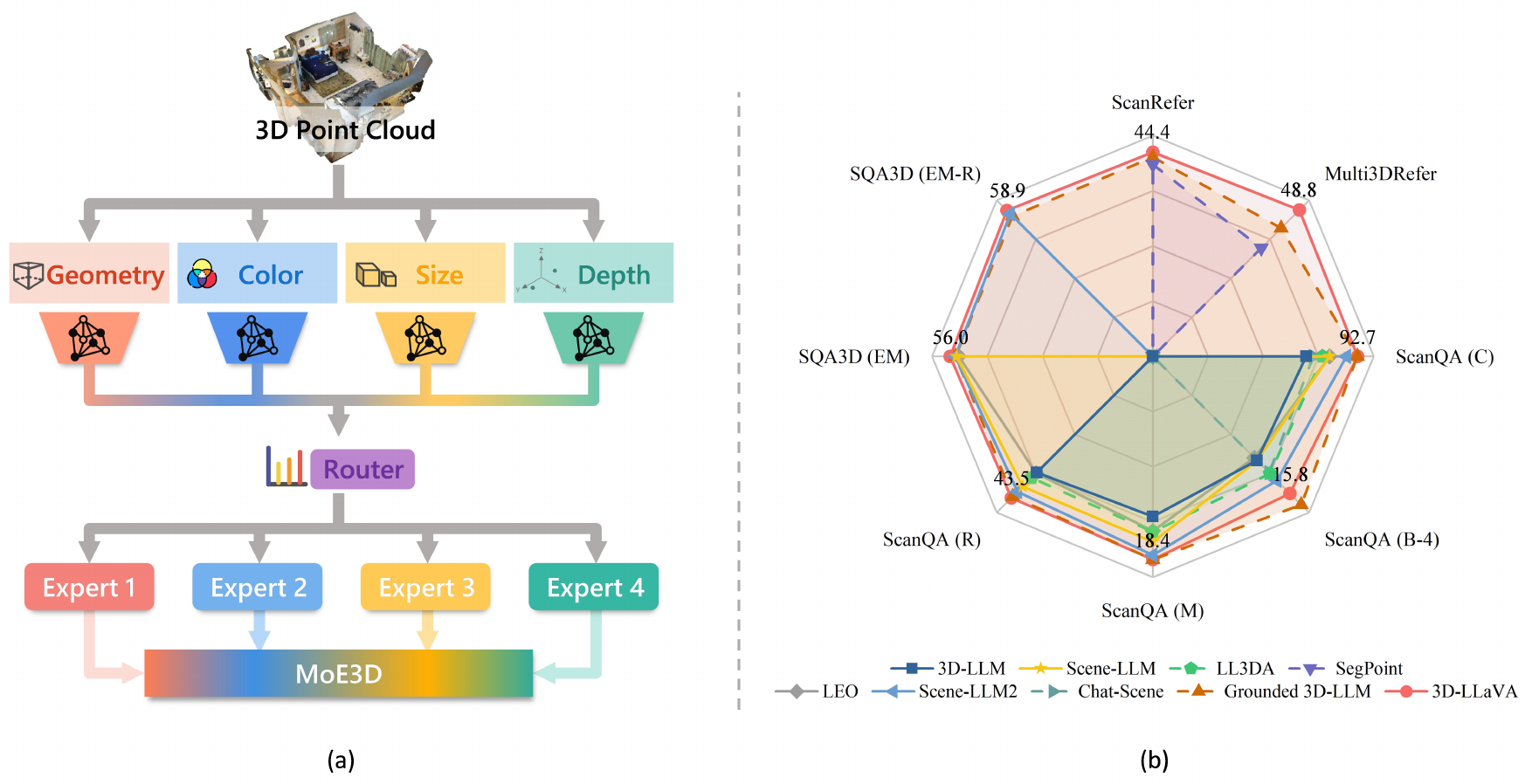}
    \caption{(a) Schematic overview of our \algorithmname~. (b) Competitive performance of our \algorithmname~against contemporary algorithms on four popular 3D tasks. } 
    \label{fig:motivation}
    \vspace{-3mm}
\end{figure*}

Multi-modal 3D understanding aims to perform comprehensive perception and reasoning tasks given multiple sensory observations~\cite{3d_survey}. It plays a pivotal role in many fields, including autonomous driving~\cite{mm3ddet_survey}, embodied perception~\cite{embodiedscan} and virtual reality~\cite{mmvr_survey}.

Previous studies put effort into designing multi-modal fusion strategies, such as early fusion, middle fusion and late fusion~\cite{Vora2019PointPaintingSF,Yoo20203DCVFGJ,Huang2020EPNetEP,Sindagi2019MVXNetMV,logonet,uniseg}. Despite the rapid development in the multi-modal field, these dense fusion networks struggle to cope with the huge heterogeneity and complexity across modalities, yielding unsatisfactory performance. Besides, conventional multi-modal fusion algorithms typically introduce extra computational burden during the fusion process, which impacts the deployment of deep models in resource-limited scenarios. It is natural to wonder if it is possible to achieve more flexible and effective fusion without incurring additional cost?  

Drawing inspirations from Mixture of Experts (MoE)~\cite{Cai2024ASO,Fedus2022ARO,Masoudnia2012MixtureOE}, we design a novel MoE-based network, dubbed \algorithmname, for multi-modal 3D understanding. Our intuition is that the model can learn to dynamically select the most suitable expert conditioned on the input scan. Our method departs from using a universal dense network for all inputs. Instead, it deploys a set of specialized "expert" networks, each adept at processing a specific modality or a mode of cross-modal interaction. A learnable gating network dynamically routes inputs to the most relevant subset of experts for multi-modal fusion. The schematic overview is depicted in Fig.~\ref{fig:motivation} (a).

Specifically, the MoE-based transformer is designed to better utilize the complementary information hidden in the visual features. Information aggregation module is put forward to further enhance the fusion performance. Top-1 gating is employed to make one expert process features within expert groups, ensuring high efficiency. We further propose a progressive pre-training strategy to better leverage the semantic and 2D prior, thus equipping the network with good initialization. As shown in Fig.~\ref{fig:motivation} (b), compared to previous competitive algorithms, our \algorithmname~consistently exhibits superior performance in four prevalent 3D understanding benchmarks.

The contributions of this paper are summarized below:

\begin{itemize}[leftmargin=*]

\item To our knowledge, we design the first MoE-based network, dubbed \algorithmname, for unified 3D perception and vision-language tasks. 

\item We design the MoE superpoint transformer (MEST) to better utilize the valuable information in superpoint features. Information aggregation module and Top-1 gating is employed to enhance fusion performance and ensure high efficiency, respectively. A progressive pre-training strategy is presented to leverage semantic and 2D prior.

\item Our \algorithmname~exhibits competitive performance in four prevalent benchmarks, which underscores the effectiveness of the proposed paradigm.

\end{itemize}
\section{Related work}
\label{sec:relatedwork}



\noindent\textbf{3D Multi-Modal Understanding.}
Traditional 3D multi-modal understanding integrates geometric and visual cues from RGB images and point clouds.
Early fusion methods such as Frustum PointNets~\cite{Qi2017FrustumPF} and PointPainting~\cite{Vora2019PointPaintingSF} enrich point clouds with projected 2D semantic features.
Mid-level approaches like 3D-CVF~\cite{Yoo20203DCVFGJ} and EPNet~\cite{Huang2020EPNetEP} perform cross-modal interaction in the feature space, while late fusion methods such as CLOCs~\cite{Pang2020CLOCsCO} combine modality-specific outputs.
Despite their effectiveness, these fixed fusion schemes struggle to adapt across diverse scenes.
Recent work incorporates large-scale pretraining and vision–language alignment into 3D multi-modal learning.
OpenScene~\cite{Peng2022OpenScene3S} transfers open-vocabulary 2D knowledge via feature back-projection.
PointLLM~\cite{Xu2023PointLLMEL} and 3D-LLM~\cite{Hong20233DLLMIT} employ LLMs for 3D reasoning through multi-view or point-level inputs, while Point-Bind~\cite{Guo2023PointBindP} aligns 3D data with ImageBind~\cite{Girdhar2023ImageBindOE}.
However, most existing systems focus on global scene reasoning and overlook complementary information hidden in the visual features critical for detailed perception.

\noindent \textbf{Mixture of Experts.}
The core idea of Mixture-of-Experts (MoE) is that a model is decomposed into multiple specialized sub-networks, called \emph{experts}, each tailored to process specific feature distributions or task domains~\cite{Cai2024ASO,Fedus2022ARO,Masoudnia2012MixtureOE}. 
MoE has achieved remarkable success in large-scale foundation models across language, vision, and multi-modal domains, 
such as Switch Transformer~\cite{Fedus2021SwitchTS}, GLaM~\cite{Du2021GLaMES}, Swin-MoE~\cite{Hwang2022TutelAM}, MoE-LLaVA~\cite{Lin2024MoELLaVAMO}, DeepSeek-V2~\cite{Shao2024DeepSeekV2AS}, and Mixtral-8x22B~\cite{Jiang2024MixtralOE}. 
Recent studies have applied MoE to vision-centric tasks, including image classification~\cite{Riquelme2021ScalingVW,Chowdhury2023PatchlevelRI}, object detection~\cite{Hwang2022TutelAM,Chen2023AdaMVMoEAM}, semantic segmentation~\cite{Jiang2024M4oEAF,Pavlitskaya2020UsingMO}, and robotic manipulation~\cite{yang2025tra}.
Nevertheless, despite its strong potential for adaptive specialization, the application of MoE in \emph{multi-modal 3D understanding} remains largely unexplored. In this work, we extend the MoE framework to 3D multi-modal understanding, enabling dynamic expert routing to multi-modality within complex 3D scenes.

\section{Methodology}
\label{sec:methodology}

\begin{figure*}[t]
    \centering
    \includegraphics[width=0.95\linewidth]{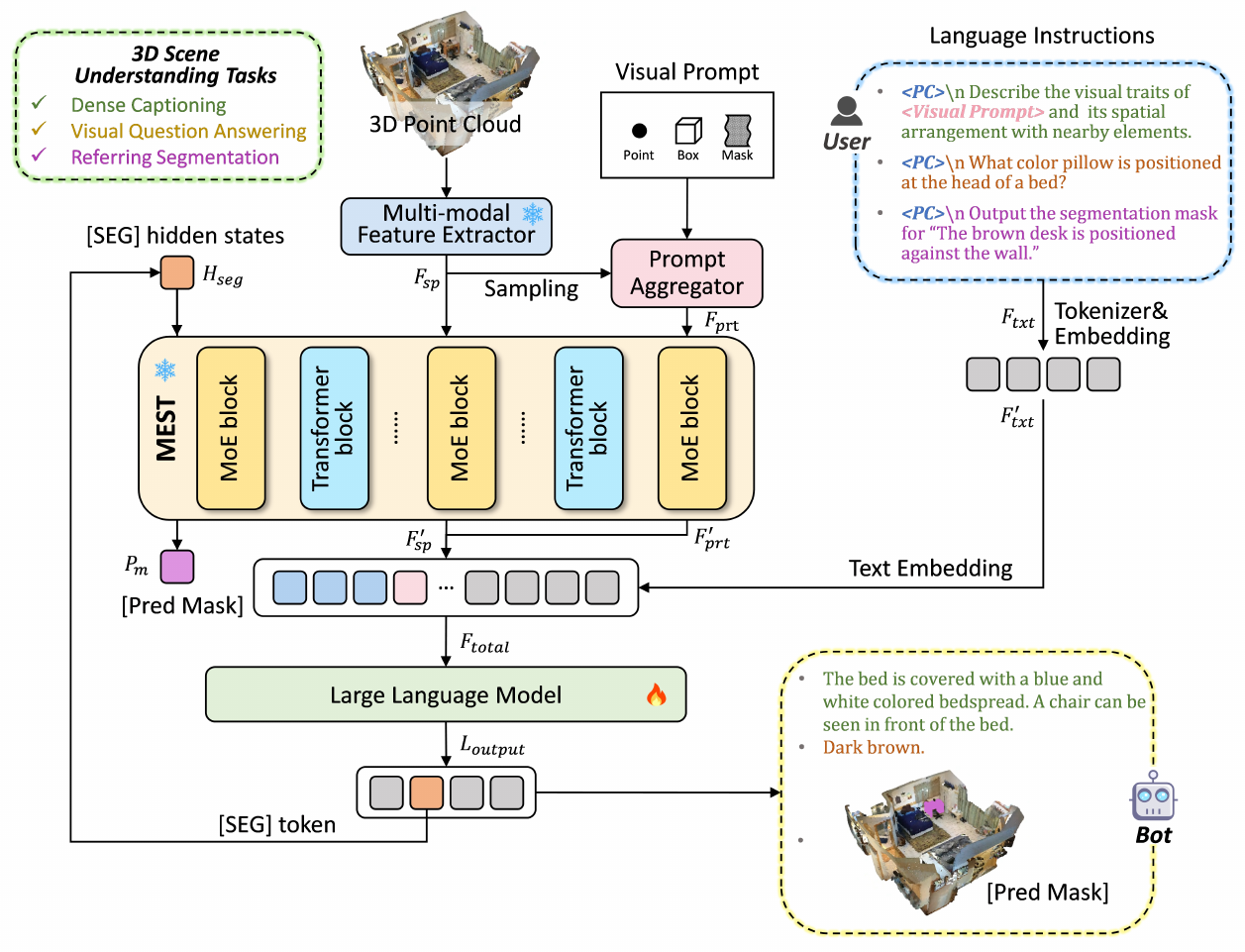}
    \caption{Framework overview of our \algorithmname~. The colored point cloud is fed to the multi-modal feature extractor and produce the visual features. Visual prompt and the sampled visual features are sent to the prompt aggregator, generating the enhanced prompt features. Visual features, together with the prompt features, are sent to the MoE superpoint transformer (MEST), generating visual tokens. The produced visual tokens and the text embedding are fed to the large language model, yielding the ultimate output. For referring segmentation, the predicted masks are subsequently produced via the MEST module. The language model is partially finetuned using LoRA~\cite{Hu2021LoRALA}.} 
    \label{fig:pipeline}
    \vspace{-3mm}
\end{figure*}

The overall framework of \algorithmname~is illustrated in Fig.~\ref{fig:pipeline}. It integrates the Mixture-of-Experts (MoE) into the multi-modal learning paradigm, where each expert is specialized for distinct modality or a mode of cross-modal interaction, thereby utilizing the complementary information hidden in the visual features and enhancing the overall capability of 3D scene understanding. Sec.~\ref{subsec:3d_feature_extractor} first presents the architecture of the multi-modal feature extractor. Sec.~\ref{subsec:moe_transformer} provides a detailed description of the core components, including MoE Superpoint Transformer (MEST) and Information Aggregation module. Finally, Sec.~\ref{subsec:training_strategy} outlines the training recipe.

\subsection{Multi-modal Feature Extractor}
\label{subsec:3d_feature_extractor}
Let $\mathcal{P} = \{\mathbf{p}_i \mid i = 1, \dots, N\}$ denote a colored point cloud consisting of $N$ points,
where each point $\mathbf{p}_i \in \mathbb{R}^{6}$ includes its 3D spatial coordinates $(x_i, y_i, z_i)$
and RGB value $(r_i, g_i, b_i)$.
Following the voxel representation adopted in~\cite{Choy20194DSC},
the point cloud $\mathcal{P}$ is discretized into a set of regular voxel grids $\mathcal{V} = \{\mathbf{v}_j\}_{j=1}^{M} \in \mathbb{R}^{M \times 6}$,
where $M$ is the number of non-empty voxels ($M < N$).
The voxels $\mathcal{V}$ are then processed using a UNet-based backbone built upon sparse 3D convolutions, producing voxel-wise feature embeddings:
\begin{align}
\mathcal{F}_v = \{\mathbf{f}^v_j\}_{j=1}^{M} \in \mathbb{R}^{M \times C}.
\end{align}
where $C$ denotes the number of channels for feature embeddings.

However, directly operating on all voxels remains computationally expensive for subsequent transformer-based reasoning.
Thus, we adopt a \textbf{superpoint-based pooling} strategy~\cite{Landrieu2017LargeScalePC}.
Specifically, given a precomputed superpoint partition of the scene,
we aggregate point features belonging to the same superpoint via average pooling to obtain:
\begin{align}
\mathcal{F}_{sp} = \{\mathbf{f}^{sp}_k\}_{k=1}^{L} \in \mathbb{R}^{L \times C},
\end{align}
where $L$ denotes the number of superpoints ($L \ll N$).

Common visual prompts include point clicks, bounding boxes, and binary masks. To extract their corresponding feature representations, the \textbf{Prompt Aggregator} applies three-nearest-neighbor (threeNN) interpolation~\cite{qi2017pointnet++} for point-click prompts, and average pooling for bounding-box and mask prompts to sample the corresponding visual prompt features $\mathcal{F}_{prt} \in \mathbb{R}^{T \times C}$ from superpoint features $\mathcal{F}_{sp}$, where $T$ is the number of prompt tokens.

\subsection{MoE Superpoint Transformer}
\label{subsec:moe_transformer}

\begin{figure*}[!t]
    \centering
    \includegraphics[width=\linewidth]{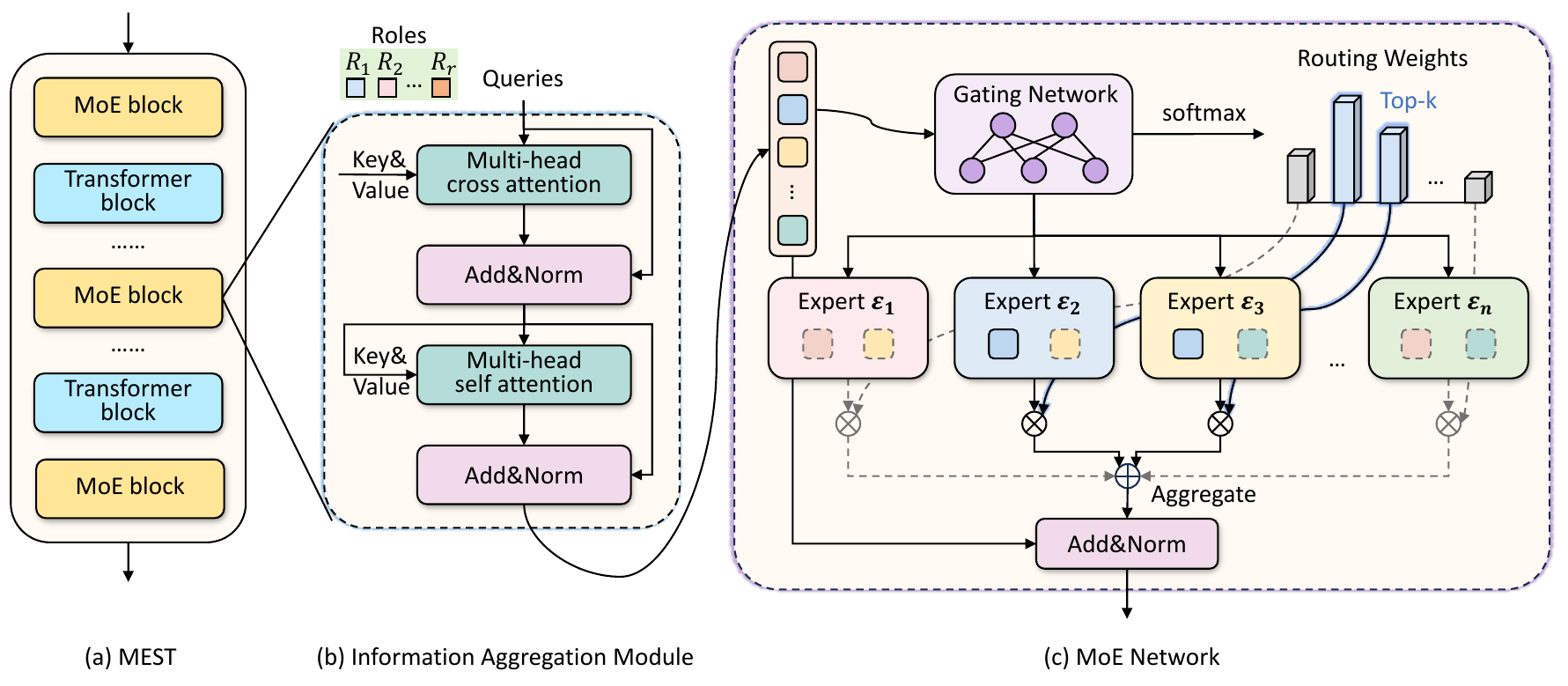}
    \vspace{-3mm}
    \caption{Overview of our MoE Superpoint Transformer. It consists of vanilla Transformer blocks and MoE blocks, where the latter are inserted in an interleaved way. Each MoE block contains four experts. In the feedforward operation, only one expert is activated by the gating network and used to process the input features, ensuring high effiency.} 
    \label{fig:MEST}
    \vspace{-3mm}
\end{figure*}

Following the success of the Mixture of Experts (MoE) approach\cite{Lin2024MoELLaVAMO}\cite{Zhong2024ConvolutionML}, we introduce MoE mechanism that dynamically routes object attributes to specialized experts, allowing adaptive selection and fusion of cues that are most informative for the current scene. 
Our MEST module, as shown in Fig.~\ref{fig:MEST} (a), replaces several layers of the standard dense Transformer with MoE blocks, consisting of two components, the Information Aggregation Module (Fig.~\ref{fig:MEST} (b)) and the MoE Network (Fig.~\ref{fig:MEST} (c)). 

\noindent \textbf{Information Aggregation Module.}
The Information Aggregation Module enhances fusion through attention-based interaction, consisting of a cross-attention layer and a self-attention layer. As illustrated in Fig.~\ref{fig:pipeline}, this unified mechanism performs three roles depending on the input: superpoint feature refinement ($\mathbf{R_1}$), prompt-based interaction ($\mathbf{R_2}$), and mask decoding for referring segmentation ($\mathbf{R_3}$).
$\mathbf{R_1}$ enhances long-range spatial dependency modeling among superpoints where cross-attention reduces to self-attention.
$\mathbf{R_2}$ allows the module to highlight spatial regions semantically relevant to the prompt, effectively linking task cues to 3D scene.
For $\mathbf{R_3}$, the LLM outputs a \texttt{[SEG]} token in its textual response $L_{\text{output}}$. 
Upon detecting this token, we extract its hidden state $H_{\text{seg}}$ and project it into the segmentation query embedding $\mathcal{F}_{seg}$. 
This query is then passed through the frozen MEST module, whose mask head produces a query-conditioned kernel that interacts with the superpoint features via a dot-product similarity, yielding the final predicted segmentation mask.

The cross-attention mechanism we designed unifies these three cases under a generalized formulation:
\begin{align}
&\mathrm{Attn}(W_i^Q, W_i^K, W_i^V) =
\mathrm{softmax}\!\left(\frac{Q_i K_i^{\!T}}{\sqrt{C}}\right)V_i, \\
&\{Q_i, K_i, V_i\}\! = \!
\begin{cases}
\{\mathcal{F}_{sp}W_1^Q,\, \mathcal{F}_{sp}W_1^K,\, \mathcal{F}_{sp}W_1^V\}, &\!\!\!\! i\!=\!R_1, \\[3pt]
\{\mathcal{F}_{prt}W_2^Q,\, \mathcal{F}_{sp}W_2^K,\, \mathcal{F}_{sp}W_2^V\}, &\!\!\!\! i\!=\!R_2, \\[3pt]
\{\mathcal{F}_{seg}W_3^Q,\, \mathcal{F}_{sp}'W_3^K,\, \mathcal{F}_{sp}'W_3^V\}, &\!\!\!\! i\!=\!R_3.
\end{cases}
\end{align}
where $\mathcal{F}_{sp}$ denotes the superpoint features, $\mathcal{F}_{prt}$ represents the visual prompt features, $\mathcal{F}_{sp}'$ is the features produced by the MEST module, and $\mathcal{F}_{seg}$ is the projected segmentation query embeddings, parameterized by $(W_i^Q, W_i^K, W_i^V)$.

After cross-attention, the updated tokens are further processed by a self-attention layer, with each block wrapped by residual connections and layer normalization.



\noindent \textbf{MoE Network.}
After capturing inter- and intra-modal dependencies through Information Aggregation Module, the features are then passed through the MoE Network, which comprises multiple feed-forward networks (FFNs), each serving as an independent expert, adept at processing a specific modality.
This design preserves the same computational cost as a standard dense Transformer, while expanding the representational capacity.

We define $\mathbf{X}_s \in \mathbb{R}^{L \times D}$ as the sequence of superpoint embeddings obtained from the Information Aggregation Module and  $\mathcal{E} = \{\varepsilon_e \mid e = 1, \dots, E\}$ as a set of parallel experts, where $D$ is the feature embedding dimension and $E$ is the number of experts. A lightweight gating network $\mathcal{G}(\cdot)$, parameterized by $W_{\mathcal{E}}$, is employed to compute the gating score $g_{s}^{\mathcal{E}}$ between the input $\mathbf{X}_s$ and all of the experts $\mathcal{E}$. Then, $g_{s}^{\mathcal{E}}$ is normalized into a probability distribution, denoted as routing weight $\mathcal{W}_\mathcal{E}^{\text{router}}  \in \mathbb{R}^{L\times E}$:
\begin{align}
& g_s^\mathcal{E} = \mathcal{G} \left (  X_s;W_{\mathcal{E}}  \right ) = W_{\mathcal{E}}^T X_s, \\
& \mathcal{W}_e^{\text{router}} = \text{softmax}\left (  g_s^{\mathcal{E} }  \right )_e = \frac{\exp(g_s^e)}{\sum_{j=1}^{E} \exp(g_s^j) }, 
\end{align}
where $\mathcal{W}_e^{\mathrm{router}} \in \mathbb{R}^{L}$ gives the probability of routing each token of $X_s$ to the $e$-th expert.
To encourage sparse expert activation, we keep only the top-$k$ entries and set the rest to zero:
\begin{align}
\tilde{\mathcal{W}}_{e}^{\mathrm{router}} =
\begin{cases}
\mathcal{W}_{e}^{\mathrm{router}}, 
& \text{if } \mathcal{W}_{e}^{\mathrm{router}} \in 
\mathrm{Top}\text{-}K\!\big(
\mathcal{W}_{\mathcal{E}}^{\mathrm{router}}, K
\big), \\[6pt]
0, & \text{otherwise.}
\end{cases}
\label{eq:topk}
\end{align}
The final MoE output for $X_s$ is obtained by the weighted aggregation of the selected experts:
\begin{align}
\mathcal{F}_s^{\text{MoE}} = \sum_{e=1}^{E} \tilde{\mathcal{W}}_{e}^{\mathrm{router}} \, \mathcal{E}_e(X_s).
\label{eq:moe_agg}
\end{align}
In our implementation, $K$ is set to 1, i.e., each token activates only its most relevant expert:
\begin{equation}
   \mathcal{F}_s^{\text{MoE}} = \tilde{\mathcal{W}}_{e^{*}}^{\mathrm{router}} \, \mathcal{E}_{e^{*}}(X_s), 
\end{equation}
where $e^*=\underset{e\in\{1,...,E\}}{\operatorname*{\operatorname*{\arg\max}}}\tilde{\mathcal{W}}_\mathcal{E}^{\mathrm{router}}$, representing the index of the most relevant expert selected for each token.

Finally, $\mathcal{F}_s^{\text{MoE}} \in \mathbb{R}^{L \times D}$ is combined with a residual connection and layer normalization.

\begin{figure}[t]
    \centering
    \includegraphics[width=0.9\linewidth]{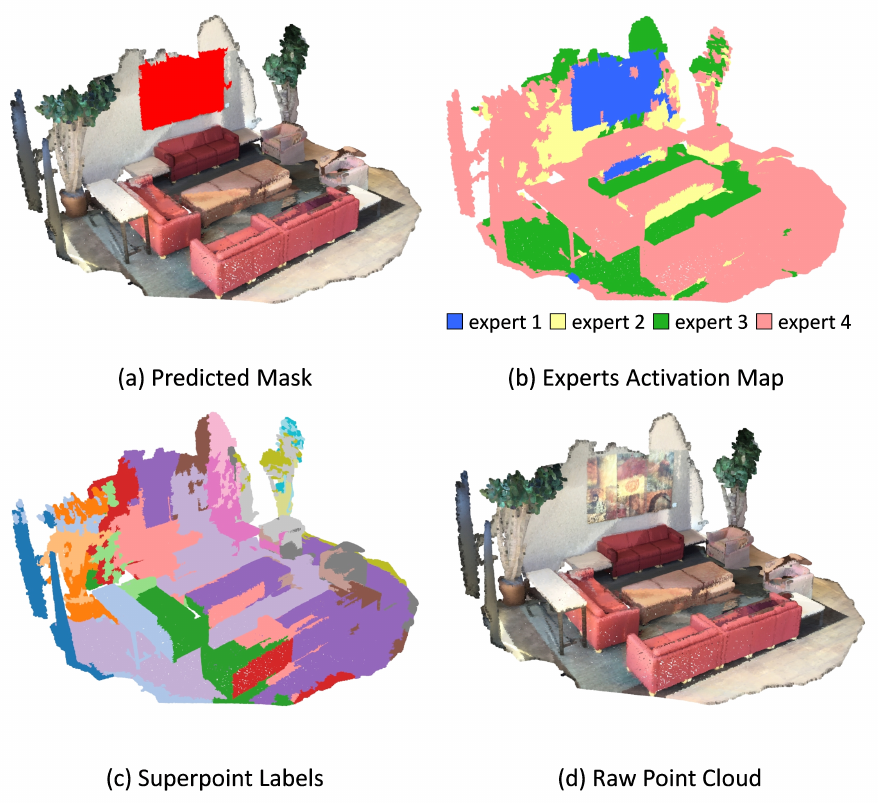}
    \caption{Visual results on the referring segmentation task. (a) Predicted mask according to the textual referring expression. (b) The four experts with different color exhibit distinct modality preferences. (c) Superpoint labels with limited boundary accuracy for training. (d) Raw point cloud of the corresponding 3D scene.} 
    \label{fig:visualize}
    \vspace{-3mm}
\end{figure}

\noindent \textbf{Visual results.}
As shown in Fig.~\ref{fig:visualize} (a), MoE3D produces accurate referring segmentation masks guided by textual instructions.
While the expert activation map reveal clear specialization among experts in Fig.~\ref{fig:visualize} (b). \textcolor{blue}{\textbf{$\text{Expert 1}$}}  predominantly activates on posters, capturing fine-grained texture and visual details. \textcolor{yellow}{\textbf{$\text{Expert 2}$}} mainly focuses on large planar wall surfaces, indicating its sensitivity to structural geometry and smooth vertical regions. \textcolor{green}{\textbf{$\text{Expert 3}$}} primarily responds to the floor area, showing a preference for horizontal and spatially continuous surfaces. \textcolor{pink}{\textbf{$\text{Expert 4}$}} is strongly activated around the red sofa area, suggesting that it specializes in color-sensitive features and high-contrast objects. These diverse activation patterns demonstrate that each expert learns complementary modalities, enabling the model to dynamically allocate specialized experts according to 3D scenes. 
Together with the superpoint annotations and raw point cloud visualization in Fig.~\ref{fig:visualize} (c) and (d), the results demonstrate that MoE3D achieves precise localization, semantic consistency, and interpretable expert behavior in complex 3D scenes.

\subsection{Training}
\label{subsec:training_strategy}
We adopt a progressive hybrid training strategy that consists of knowledge transfer learning for multi-modal feature extraction, pretraining of the MEST module, and unified instruction tuning.
This approach enables the model to better leverage semantic and 2D priors, thus equipping the network with good initialization and effectively adapting to diverse 3D tasks.

\noindent \textbf{2D-to-3D Knowledge Transfer} aims to bridge the modality gap between images and 3D point clouds, enabling the network to learn discriminative 3D representations without relying on large-scale annotated 3D datasets.
Following OpenScene\cite{Peng2022OpenScene3S}, we first obtain multi-view images $\mathcal{I}$ from 3D point cloud $\mathcal{P}$, then extract dense per-pixel embeddings from $\mathcal{I}$ using a frozen 2D segmentation model $E^{\text{2D}}$, such as OpenSeg\cite{Ghiasi2021OpenVocabularyIS} or LSeg\cite{Li2022LanguagedrivenSS}. According to the correspondence between 2D and 3D, we backproject the 2D features onto the 3D points and fuse the multi-view embeddings to obtain semantic-enriched features $\mathcal{F}^{\mathrm{2D\_3D}}$. Formally, this process can be expressed as:
\begin{align}
\mathcal{F}^{\mathrm{2D\_3D}} =
\Omega\!\left( 
\left\{ 
\Theta\!\left(E^{\text{2D}}(\mathcal{I}_i)\right)
\right\}_{i=1}^{\nu}
\right),
\end{align}
where $\mathcal{I}_i \in \mathbb{R}^{H \times W \times 3} $, $\nu$ is the number of views, $\Theta$ denotes the reprojection operation from 2D pixels to 3D points, and $\Omega$ represents the multi-view feature fusion process.

To ensure the 3D point cloud features $\mathcal{F}^{\text{3D}}$ extracted by the 3D feature extractor $E^{\text{3D}}$ aligned with $\mathcal{F}^{\mathrm{2D\_3D}}$, we employ a cosine similarity loss function defined as:
\begin{align}
& \mathcal{L}_{\text{align}} = 1 - \cos(\mathcal{F}^{\mathrm{2D\_3D}}, \mathcal{F}^{\text{3D}}), \\
& \mathcal{F}^{3D} = E^{\text{3D}}(\mathcal{P}),
\end{align}
where SPConvUnet\cite{Chen2024SSTNetSS} is adopted as $E^{\text{3D}}$ in practise.


\noindent \textbf{Pre-training MoE Superpoint Transformer.} 
We conduct hybrid supervised pretraining on the ScanNet200 dataset to optimize the proposed MEST module. Specifically, the pretraining objective combines the  MoE regularization and the segmentation task.

To enhance the stability of the expert routing process, we incorporate a router z-loss following ST-MoE~\cite{Zoph2022STMoEDS}, which discourages the gating network from producing excessively large logits.
Formally, the router z-loss is defined as:
\begin{equation}
\mathcal{L}_{z} = \frac{1}{S} \sum_{k=1}^{S} \left( \log \sum_{i=1}^{E} e^{g_i^{(k)}} \right)^{2},
\end{equation}
where $S$ denotes the total number of tokens, $E$ is the number of experts, and $g \in \mathbb{R}^{S \times E}$ represents the router logits.

For segmentation tasks, following Oneformer3D\cite{Kolodiazhnyi2023OneFormer3DOT}, classification errors are penalized with a cross-entropy loss \(\mathcal{L}_{cls}\) . Besides, the superpoint mask loss is computed by a binary cross-entropy \(\mathcal{L}_{bce}\) and a Dice loss \(\mathcal{L}_{dice}\). The semantic loss \(\mathcal{L}_{bce}\) is also defined as a binary cross-entropy. The total segmentation loss is formulated as:
\begin{equation}
\mathcal{L}_{seg} = \lambda_{\text{cls}} \mathcal{L}_{\text{cls}} + \lambda_{\text{bce}} \mathcal{L}_{\text{bce}} + \lambda_{\text{dice}} \mathcal{L}_{\text{dice}} + \lambda_{sem} \mathcal{L}_{sem}.
\end{equation}

The overall objective for pre-training our MEST module is given by:
\begin{equation}
\mathcal{L}_{inst} = \mathcal{L}_{\text{seg}} + \lambda_{z} \mathcal{L}_{z}.
\end{equation}

\noindent \textbf{Unified Instruction Tuning.} 
We unify various 3D tasks such as referring segmentation, visual question answering, and dense captioning into conditional generation problems, adapting the model to different task requirements through instruction tuning. 
Specifically, we jointly train the model with task-specific instructions and prompts, which guide the model in generating corresponding outputs.

The model outputs consist of text generation and superpoint mask prediction. 
The textual response is generated by the LLM and optimized with a cross-entropy loss, while the mask prediction yields binary superpoint masks, supervised by a binary cross-entropy loss and a Dice loss. The training loss is formulated as:
\begin{equation}
\mathcal{L}_{ft} = \mathcal{L}_{text} + \lambda_{m}\times \mathcal{L}_{mask}.
\end{equation}

\section{Experiments}
\label{sec:experiment}

\noindent \textbf{Benchmarks.} 
During the first and second pre-training stage, we leverage ScanNet200~\cite{Rozenberszki2022LanguageGroundedI3}, which exhibits a long-tailed distribution of labels. 
For instruction tuning, we curate a unified training corpus by combining diverse vision-language datasets, including referring segmentation benchmarks (ScanRefer~\cite{Chen2019ScanRefer3O}, Nr3D~\cite{Achlioptas2020ReferIt3DNL}, Multi3DRefer~\cite{Zhang2023Multi3DReferGT}), 3D question–answering datasets (ScanQA~\cite{azuma2022scanqa}, SQA3D~\cite{Ma2022SQA3DSQ}), and the dense captioning dataset (Scan2Cap~\cite{chen2021scan2cap} and Nr3D~\cite{Achlioptas2020ReferIt3DNL} reused as complementary caption data~\cite{Huang2023ChatSceneB3}), which provides richer linguistic descriptions to enhance the instruction-following capability of the model.
We evaluate our model on four benchmarks, Multi3DRefer~\cite{Zhang2023Multi3DReferGT}, ScanRefer~\cite{Chen2019ScanRefer3O}, ScanQA~\cite{azuma2022scanqa}, and SQA3D~\cite{Ma2022SQA3DSQ}.

\noindent \textbf{Evaluation metrics.}
We assess the quality of generated text outputs for ScanQA~\cite{azuma2022scanqa} following standard evaluation protocols, using CiDEr (C), BLEU-4 (B-4), METEOR (M), and Rouge-L (R) metrics. In contrast to the conventional ScanQA~\cite{azuma2022scanqa} setting, the SQA3D~\cite{Ma2022SQA3DSQ} dataset provides explicit ground-truth answers for each question, thus we employ Exact Match (EM) and its refined variant EM-R to measure accuracy. For referring segmentation tasks, the evaluation is conducted using the mean Intersection over Union (mIoU) metric.

\noindent \textbf{Implementation details.}
Following \cite{Landrieu2017LargeScalePC} , we apply a graph-based superpoint clustering method on ScanNet200~\cite{Rozenberszki2022LanguageGroundedI3}, with the voxel size set to 2~cm. 
Within the MEST module, we replace the 1st, 3rd, and 6th layers with MoE blocks. By default, the number of experts is set to 4, and a top-1 gating strategy is adopted for token routing, ensuring that the computational cost (FLOPs) per token remains approximately constant.
During training, we adopt LoRA\cite{Hu2021LoRALA} to the LLM(Vicuna-1.5-7B\cite{chiang2023vicuna}) and keep the multi-modal feature extractor, the MEST module and the main body of LLM frozen. 
All experiments are conducted on 8$\times$NVIDIA RTX 4090 GPUs. We use the AdamW optimizer with a Cosine Annealing learning rate schedule, initializing the learning rate at $2\times10^{-4}$. The batch size is set to 2 per GPU, and model parameters are updated after accumulating gradients for 8 steps.

\subsection{Quantitative comparison}
The detailed performance comparison between our \algorithmname~and contemporary algorithms is summarized in Table~\ref{3dllm_comparison}. The compared methods can be divided into three categories: specialist models, finetuned 3D LLMs, and 3D LLMs. 
Notably, unlike the aforementioned methods, our work introduces the MoE (Mixture-of-Experts) mechanism into 3D scene understanding for the first time, leveraging the complementary information of different modalities and enabling adaptive understanding of complex 3D scenes through dynamic expert activation. This mechanism allows the model to handle both point-level perception tasks (e.g., referring segmentation) and language generation tasks (e.g., 3D question answering and scene description) within a unified framework, achieving a better balance between task specialization and cross-task generalization.

\begin{table*}[t]
\caption{Performance comparison among state-of-the-art methods on four 3D understanding benchmarks. “Specialist Model” are methods developed for a single task (e.g., 3D QA or referring segmentation). “Finetuned 3D LMM” refers to models pretrained on multiple tasks and then fine-tuned on each benchmark before evaluation (marked with “*”). “3D LMM” indicates models trained jointly on multiple tasks. “PC” denotes point clouds and “I” denotes multi-view images. Results of LEO\cite{Huang2023AnEG} on ScanQA are under a different setting (using ground-truth object information) and thus not directly comparable. The best algorithm is in bold and the second is underlined.}
\centering
\scriptsize
\resizebox{\textwidth}{!}{%
\begin{tabular}{lcccccccccc}
\toprule
\textbf{Method} & \textbf{Modality} & \textbf{ScanRefer(val)} & \textbf{Multi3DRefer(val)} & \multicolumn{4}{c}{\textbf{ScanQA(val)}} & \multicolumn{2}{c}{\textbf{SQA3D(test)}} \\
\cmidrule(lr){3-3} \cmidrule(lr){4-4} \cmidrule(lr){5-8} \cmidrule(lr){9-10}
 &  & mIoU↑ & mIoU↑ & C↑ & B-4↑ & M↑ & R↑ & EM↑ & EM-R↑ \\
\midrule
\multicolumn{10}{l}{\textit{Specialist Models:}} \\
ScanQA\cite{azuma2022scanqa} & PC & - & - & 64.9 & 10.1 & 13.1 & 33.3 & 46.6 & - \\
3D-VLP\cite{jin2023context} & PC & - & - & 67.0 & 11.2 & 13.5 & 34.5 & 48.5 & - \\
3D-VisTA\cite{zhu20233d} & PC & - & - & 69.6 & 10.4 & 13.9 & \textbf{45.7} & 48.5 & - \\
Scan2Cap\cite{chen2021scan2cap} & PC & - & - & - & - & - & - & 41.0 & - \\
MORE\cite{jiao2022more} & PC & - & - & - & - & - & - & - & - \\
SpaCap3D\cite{wang2022spatiality} & PC & - & - & - & - & - & - & - & - \\
D3Net\cite{Chen2021D3NetAU} & PC & - & - & - & - & - & - & - & - \\
UniT3D\cite{chen2023unit3d} & PC & - & - & - & - & - & - & - & - \\
3DJCG\cite{cai20223djcg} & PC & - & - & - & - & - & - & - & - \\
Vote2Cap-DETR\cite{Chen2023EndtoEnd3D} & PC & - & - & - & - & - & - & - & - \\
TGNN\cite{Huang2021TextGuidedGN} & PC & 27.8 & - & - & - & - & - & - & - \\
M3DRef-CLIP\cite{Zhang2023Multi3DReferGT} & PC & 35.7 & 32.6 & - & - & - & - & - & - \\
X-RcfSeg3D\cite{Qian2024XRefSeg3DER} & PC & 29.9 & - & - & - & - & - & - & - \\
3D-STMN\cite{Wu20233DSTMNDS} & PC & 39.5 & - & - & - & - & - & - & - \\
\midrule
\multicolumn{10}{l}{\textit{Finetuned 3D LMMs:}} \\
3D-LLM\cite{Hong20233DLLMIT} & PC+I & - & - & 69.4 & 12.0 & 14.5 & 35.7 & - & - \\
Scene-LLM*\cite{Fu2024SceneLLMEL} & PC+I & - & - & 80.0 & 12.0 & 16.8 & 40.0 & 54.2 & - \\
LL3DA*\cite{Chen2023LL3DAVI} & PC & - & - & 76.8 & 13.5 & 15.9 & 37.3 & - & - \\
SegPoint*\cite{He2024SegPointSA} & PC & 41.7 & 36.1 & - & - & - & - & - & - \\
\midrule
\multicolumn{10}{l}{\textit{3D LMMs:}} \\
LEO\cite{Huang2023AnEG} & PC+I & - & -  & \textcolor{gray}{101.4} & \textcolor{gray}{13.2} & \textcolor{gray}{20.0} & \textcolor{gray}{49.2} & 50.0 & 52.4 \\
Scene-LLM\cite{Fu2024SceneLLMEL} & PC+I & - & - & 80.0 & 11.7 & 15.8 & 35.9 & 53.6 & - \\
Chat-Scene\cite{Huang2023ChatSceneB3} & PC+I & - & - & 87.7 & 14.3 & 18.0 & 41.6 & \underline{54.6} & \underline{57.5} \\
Grounded 3D-LLM\cite{Chen2024Grounded3W} & PC & - & - & 72.7 & 13.4 & - & - & - & - \\
3D-LLaVA\cite{Deng20253DLLaVATG} & PC & \underline{43.3} & \underline{42.7} & \underline{92.6} & \textbf{17.1} & \textbf{18.4} & 43.1 & 54.5 & 56.6 \\
\rowcolor{violet!10}
\textbf{MoE3D(ours)} & PC & \textbf{44.4} & \textbf{48.8} & \textbf{92.7} & \underline{15.8} & \textbf{18.4} & \underline{43.5} & \textbf{56.0} & \textbf{58.9} \\
\bottomrule
\end{tabular}%
}
\label{3dllm_comparison}
\end{table*}

\noindent \textbf{3D Referring Segmentation} evaluates a model’s ability to interpret natural-language expressions and localize the described objects in a 3D scene by predicting instance-level masks.
We assess performance on two benchmarks: ScanRefer~\cite{Chen2019ScanRefer3O}, which contains single-object references, and Multi3DRefer~\cite{Zhang2023Multi3DReferGT}, where one expression may refer to one, multiple, or no objects.
Following the protocol in~\cite{He2024SegPointSA}, masks corresponding to multiple referenced objects are merged into a single region for evaluation, and empty masks are given when no target exists.
As shown in Table~\ref{3dllm_comparison}, our MoE3D achieves state-of-the-art performance on both datasets, reaching $44.4\%$ mIoU on ScanRefer and $48.8\%$ mIoU on Multi3DRefer—improving over the prior best SegPoint by $+1.1\%$ and $+6.1\%$ mIoU, respectively.

\noindent \textbf{3D Question Answering}  aims to understand and reason about spatial relationships, object attributes, and semantic content within 3D scenes based on natural language questions, thereby generating accurate textual answers.
We conduct evaluations on two benchmarks: ScanQA~\cite{azuma2022scanqa} for standard scene-level reasoning and SQA3D~\cite{Ma2022SQA3DSQ} for situated, context-grounded question answering. 
As shown in Table~\ref{3dllm_comparison}, our MoE3D achieves state-of-the-art performance across multiple benchmarks. 
It attains the best scores on SQA3D ($56.0\%$ EM and $58.9\%$ EM-R), 
surpassing previous methods by $+1.4\%$ EM and $+1.4\%$ EM-R. 
On ScanQA, MoE3D achieves $92.7$ CiDEr, ranking second while maintaining competitive results across BLEU-4, METEOR, and Rouge-L metrics.

\subsection{Ablation study}


\noindent \textbf{The Number of Experts.}
Table~\ref{tab:num_experts} examines the effect of varying the number of experts~$e$. 
Performance improves consistently when increasing $e$ from 1 to 4, reaching the best mIoU on both ScanRefer and Multi3DRefer, indicating that moderate expert diversity enhances the modeling of heterogeneous modality priors. 
Larger expert counts, however, yield diminishing returns due to routing instability and fragmented token assignments. 
Across all settings, GFLOPs and latency remain nearly constant, as Top-$K$ sparse routing activates only one expert per token. 
All FLOP measurements are obtained using \texttt{torch.profiler} with operator-level FLOP counting enabled.


\begin{table}[t]
\centering
\caption{
Ablation on the number of experts $e$ (Top-$K$=1, \#replaced\_layers = 3). 
We report performance on ScanRefer~\cite{Chen2019ScanRefer3O} and Multi3DRefer~\cite{Zhang2023Multi3DReferGT}, both using mIoU as the metric. Additionally report model size, computational cost, and inference latency. 
Our default setting is highlighted with \colorbox{violet!10}{light violet}.
}
\label{tab:num_experts}
\resizebox{\columnwidth}{!}{%
\begin{tabular}{lcccccc}
\toprule
\multirow{2}{*}{$e$} &
\multirow{2}{*}{\#Params (B)} &
\multirow{2}{*}{GFLOPs} &
\multirow{2}{*}{\begin{tabular}[c]{@{}c@{}}Inference\\ Latency (ms)\end{tabular}} &
\multicolumn{1}{c}{ScanRefer} &
\multicolumn{1}{c}{Multi3DRefer} \\
\cmidrule(r){5-5}
\cmidrule(l){6-6}
 &  &  &  & mIoU↑ & mIoU↑ \\
\midrule
1  & 6.79 & 16.75 & 152.71 & 43.5 & 42.8 \\
2  & 6.80 & 16.75 & 148.59 & 42.1 & 47.2 \\
\rowcolor{violet!10}
4  & 6.81 & 16.77 & 144.08 & \textbf{44.4} & \textbf{48.8} \\
6  & 6.82 & 16.74 & 151.61 & 42.3 & 47.2 \\
8  & 6.83 & 16.73 & 151.63 & 41.5 & 46.2 \\
\bottomrule
\end{tabular}%
}
\end{table}

\noindent \textbf{MoE v.s. FFN.}
Table~\ref{tab:num_moe_layers} compares the performance of the standard Transformer with feed-forward networks (FFN-only) and our variants where a subset of FFN layers are replaced with Mixture-of-Experts (MoE) layers. All models are trained under identical settings to ensure a fair comparison. We observe that partially integrating MoE layers (i.e., replacing 3 out of 6 Transformer layers) yields the highest accuracy on the SQA3D benchmark, achieving an EM score of $56.0$ and an EM-R score of $58.9$. This demonstrates that introducing expert specialization enhances the model’s representational capacity and enables more adaptive multi-modal reasoning. However, replacing all FFN layers with MoE layers (6/6) slightly reduces performance, likely due to routing redundancy and over-fragmentation of expert utilization across layers. In contrast, using only a small proportion of MoE layers (1/6) does not provide sufficient expert diversity. Importantly, the computational cost is comparable to the baseline, indicating that our sparse routing mechanism achieves improved performance without sacrificing efficiency.

\begin{table}[t]
\centering
\caption{ Ablation study comparing standard FFN layers and MoE layers in the Transformer (Top-$K$=1, $e$=4). Evaluation is conducted on the SQA3D~\cite{Ma2022SQA3DSQ} benchmark using EM and EM-R metrics. 
}
\label{tab:num_moe_layers}
\resizebox{\columnwidth}{!}{%
\begin{tabular}{lccccc}
\toprule
\multirow{2}{*}{\#Replaced layers} &
\multirow{2}{*}{\#Params (B)} &
\multirow{2}{*}{GFLOPs} &
\multirow{2}{*}{\begin{tabular}[c]{@{}c@{}}Inference\\ Latency (ms)\end{tabular}} &
\multicolumn{2}{c}{SQA3D} \\
\cmidrule(lr){5-6}
 &  &  &  & EM↑ & EM-R↑ \\
\midrule
0 (FFN-only) & 6.79 & 16.90 & 144.95 & 55.5 & 58.4 \\
1            & 6.80 & 16.82 & 151.82 & 54.6 & 57.3 \\
\rowcolor{violet!10} 
3            & 6.81 & 16.77 & 144.08 & \textbf{56.0} & \textbf{58.9} \\
6            & 6.82 & 16.62 & 154.66 & 55.3 & 57.8 \\
\bottomrule
\end{tabular}%
}
\end{table}

\noindent \textbf{MoE positions.}
Table~\ref{tab:moe_position} examines the influence of inserting MoE blocks at different depths within the 6-layer Transformer. 
We compare four placement strategies: shallow, middle, deep, and interleaved integration. 
Among them, the \textit{interleaved} configuration, where MoE layers are distributed across shallow, intermediate, and deep stages ([1,3,6]), achieves the best overall results. 
This suggests that balancing expert specialization throughout the network enables complementary learning across various modalities. 
By contrast, stacking MoE layers only at the early or middle stages limits their access to abstract semantics, while placing them only at the late stages restricts their interaction with fine-grained spatial cues.

\begin{table}[t]
\centering
\caption{
Ablation on MoE layers positions within the 6-layer Transformer. 
Each model replaces three FFN layers with MoE blocks (Top-$K$=1, $e$=4). 
Performance is evaluated on ScanRefer~\cite{Chen2019ScanRefer3O} and Multi3DRefer~\cite{Zhang2023Multi3DReferGT} using mIoU, and on SQA3D~\cite{Ma2022SQA3DSQ} using EM-R.
}
\label{tab:moe_position}
\resizebox{\columnwidth}{!}{%
\begin{tabular}{lcccc}
\toprule
\multirow{2}{*}{Configuration} &
\multirow{2}{*}{\begin{tabular}[c]{@{}c@{}}MoE\\ Layer Index\end{tabular}} &
\multicolumn{1}{c}{ScanRefer} &
\multicolumn{1}{c}{Multi3DRefer} &
\multicolumn{1}{c}{SQA3D} \\
\cmidrule(r){3-3}
\cmidrule(lr){4-4}
\cmidrule(l){5-5}
 &  & mIoU↑ & mIoU↑ & EM-R↑ \\
\midrule
Shallow       & [1,2,3]   & 42.3 & 47.1 & 57.7 \\
Middle        & [2,3,4]   & 41.7 & 46.3 & 57.8 \\
Deep          & [4,5,6]   & 39.5 & 45.3 & 58.3 \\
\rowcolor{violet!10}
Interleaved   & [1,3,6]   & \textbf{44.4} & \textbf{48.8} & \textbf{58.9} \\
\bottomrule
\end{tabular}%
}
\end{table}

\noindent \textbf{Multi-modal fusion.}
Table~\ref{tab:multi_modal_fusion} compares different fusion strategies.
Early fusion directly combines raw color and position inputs, preserving pixel–point alignment and thus delivering strong performance.
Middle fusion first encodes RGB and point cloud streams separately and then fuses their intermediate features. However, the weak cross-modal interaction in this stage leads to a notable performance drop.
Late fusion, which merges modalities only at the prediction stage, performs significantly worse and is therefore omitted.
In contrast, our method achieves the best results on both ScanQA~\cite{azuma2022scanqa} and SQA3D~\cite{Ma2022SQA3DSQ}, demonstrating that dynamically routed expert specialization enables more flexible and adaptive multi-modal integration in complex 3D scenes.

\begin{table}[t]
\centering
\caption{
Ablation on multi-modal fusion strategies. We compare early and middle fusion with our method. Late fusion performs significantly worse and is omitted from comparison.
Performance is evaluated on the ScanQA~\cite{azuma2022scanqa} benchmark using four metrics (C, B-4, M, R) and on SQA3D~\cite{Ma2022SQA3DSQ} using EM.
}
\label{tab:multi_modal_fusion}
\resizebox{\columnwidth}{!}{%
\begin{tabular}{lccccc}
\toprule
\multirow{2}{*}{Fusion Strategy} &
\multicolumn{4}{c}{ScanQA} & SQA3D \\
\cmidrule(lr){2-5} \cmidrule(lr){6-6}
 & C↑ & B-4↑ & M↑ & R↑ & EM↑ \\
\midrule
Early      & 92.6 & 17.1 & 18.4 & 43.1 & 54.5 \\
Middle     & 80.0 & 11.7 & 15.8 & 35.9 & 53.6 \\
\rowcolor{violet!10} Ours & \textbf{92.7} & \textbf{15.8} & \textbf{18.4} & \textbf{43.5} & \textbf{56.0} \\
\bottomrule
\end{tabular}%
}
\end{table}
\section{Conclusion}
\label{sec:conclusion}

In this paper, we design the MoE-based network for multi-modal 3D understanding. Compared with classical multi-modal fusion methods, our \algorithmname~ achieves better performance with higher efficiency. On four popular 3D benchmarks, our \algorithmname~obtains state-of-the-art performance. We hope our study can inspire more architectural redesigns for multi-modal perception and vision-language tasks.
{
    \small
    \bibliographystyle{ieeenat_fullname}
    \bibliography{main}
}

\appendix
\clearpage
\setcounter{page}{1}
\maketitlesupplementary

\section{Additional Implementation Details}

\subsection{Datasets}
We present the detailed statistics for training and testing data in Table~\ref{tab:dataset_stats}.
Following previous work~\cite{Huang2023Chat3DVB, Huang2023AnEG}, we report the model performance on the validation set for ScanRefer~\cite{Chen2019ScanRefer3O}, Multi3DRefer~\cite{Zhang2023Multi3DReferGT}, Scan2Cap~\cite{chen2021scan2cap}, ScanQA~\cite{azuma2022scanqa}, and the test set for SQA3D~\cite{Ma2022SQA3DSQ}. 

\paragraph{ScanRefer~\cite{Chen2019ScanRefer3O}.}
ScanRefer is a 3D referring segmentation benchmark that pairs natural language expressions with target objects in 3D indoor scenes from ScanNet~\cite{Chen2019ScanRefer3O}.  We follow the official split, which uses 36{,} 665 language samples for training and 9{,} 508 language samples for validation, with a total dataset size of 37 K. 
For evaluation, ScanRefer reports the mean Intersection over Union (mIoU).
This benchmark measures the capability of the model to correctly localize the object referenced by the input sentence within cluttered, real-world 3D environments.

\paragraph{Nr3D~\cite{Achlioptas2020ReferIt3DNL}.}
Nr3D (Natural ReferIt3D) focuses on 3D referring segmentation over individual objects in indoor scenarios.
Each query is a human-written utterance that uniquely identifies an object among distractors. 
The dataset includes 28{,} 716 training language examples and the overall dataset size is 29 K.

\paragraph{Nr3D* (Caption)~\cite{Achlioptas2020ReferIt3DNL}.}
Nr3D* is the dense captioning counterpart of Nr3D, where each object instance is accompanied by a descriptive caption. 
Similar to Nr3D, the dataset contains 28{,} 716 language samples in the training split with a total dataset size of 29 K. 

\paragraph{Multi3DRefer~\cite{Zhang2023Multi3DReferGT}.}
Multi3DRefer extends referring segmentation to many-object settings where a single text query may refer to zero, one or multiple objects in the scene. This significantly increases the linguistic and visual complexity. 
The dataset contains 43{,} 838 training language samples and 11{,} 120 validation language samples, totaling 44 K. 
The evaluation metric used is mIoU. We follow the official settings of this benchmark to ensure fair evaluation.

\paragraph{ScanQA~\cite{azuma2022scanqa}.}
ScanQA is a 3D visual question answering benchmark built upon ScanNet. 
Each sample includes a natural language question that requires spatial reasoning or semantic understanding, along with a free-form text answer. 
The dataset provides 26{,} 515 training samples and 4{,} 675 validation samples, with a total size of 30 K. 
Evaluation metrics include BLEU-4(B-4), METEOR(M), ROUGE(R) and CIDEr(C) to measure the similarity of generated answers with different question types.

\paragraph{SQA3D~\cite{Ma2022SQA3DSQ}.}
SQA3D is a situated 3D question answering dataset where each question is grounded in a specific object or region within the 3D scene. Unlike ScanQA, SQA3D requires a definite answer. Therefore, we leverage exact match accuracy (EM) and the refined version (EM-R) as the metric. 
The test split contains 26{,} 623 samples with 3{,} 519 annotated supporting facts, and its official dataset size is 89~K.

\paragraph{Scan2Cap~\cite{chen2021scan2cap}.}
Scan2Cap is a 3D dense captioning dataset designed to describe object-level semantics in 3D scenes that utilizes texts from the ScanRefer~\cite{Chen2019ScanRefer3O} dataset. 
To evaluate the quality of the generated captions, we adopt standard text similarity metrics, including BLEU-4 (B-4), METEOR (M), ROUGE-L (R), and CIDEr (C).
These metrics are computed under different Intersection over Union (IoU) thresholds, allowing to assess caption accuracy conditioned on the spatial alignment between the predicted region and the ground-truth object.
The dataset includes 36{,} 665 training samples and 2{,} 007 validation samples, with a total size of 37 K. 
Scan2Cap measures the ability of models to generate expressive natural descriptions grounded in 3D geometry.

\begin{table}[t]
\centering
\caption{
Statistics of datasets used in our experiments, including the number of language samples in each split and the total size of each dataset.
}
\label{tab:dataset_stats}
\resizebox{\columnwidth}{!}{%
\begin{tabular}{l l c c c}
\toprule
\textbf{Dataset} & \textbf{Task} & \textbf{Train} & \textbf{Val/Test} & \textbf{Size} \\
\midrule
ScanRefer (val)       & referring segmentation        & 36,665 & 9,508  & 37 K \\
Nr3D            & referring segmentation        & 28,716 & --     & 29 K \\
Multi3DRefer(val)    & referring segmentation        & 43,838 & 11,120 & 44 K \\
ScanQA (val)          & visual question answering     & 26,515 & 4,675  & 30 K \\
SQA3D (test)    & visual question answering     & 26,623 & 3,519  & 89 K \\
Scan2Cap (val)       & dense captioning              & 36,665 & 2,007  & 37 K \\
Nr3D*           & dense captioning              & 28,716 & --     & 29 K \\
\rowcolor{violet!10}
Total           & --                            & 227,738 & 30,829 & 295 K \\
\bottomrule
\end{tabular}%
}
\end{table}

\subsection{Model Architecture}
\noindent \textbf{Multi-modal Feature Extractor.}
Our multi-modal feature extractor is instantiated as a sparse 3D U-Net backbone following prior work~\cite{Lai2023MaskAttentionFreeTF, Lai2023LISARS, team2023gemini,Chen2024SSTNetSS}.
It consists of a five-stage encoder–decoder architecture which progressively increases channel widths in the encoder and achieves symmetric reductions in the decoder.
Starting from 32 channels in the first stage, the feature dimensionality increases by 32 at each subsequent encoder level, and decreases accordingly during decoding, resulting in a channel configuration of "32-64-96-128-160-128-96-64-32".
A final linear projection maps the U-Net output to a 256-dimensional embedding, which is used as the input of our MoE Superpoint Transformer.

\noindent \textbf{MoE Superpoint Transformer.}
Our MEST module consists of 6 blocks. Among them, the 1st, 3rd, and 6th blocks are implemented as MoE blocks, while the remaining is the standard dense Transformer blocks. Each expert in our MoE block is a two-layer MLP with LayerNorm, GELU activation, and dropout regularization. The number of experts is set to 4, and the top-1 gating strategy is adopted for token routing.
Each block operates on a 256-dimensional hidden representation, while the feed-forward networks expand the dimension to 1024.
Both the cross-attention and self-attention blocks adopt 8 heads.
For prediction, we employ lightweight heads:
the classification head is a single linear layer that outputs 199 channels (198 object categories plus one background class), and
the mask head, also implemented as a linear projection, produces a 1024-dimensional mask embedding.

\subsection{Training Configuration}
In this section, we provide a detailed explanation of our training procedure. We implement our framework using PyTorch~\cite{Paszke2019PyTorchAI}.
For pretraining the multi-modal feature extractor, we adopt the AdamW optimizer with an initial learning rate of $1\times 10^{-4}$, and train the model for 20 epochs. For 2D multi-view feature extraction, we employ LSeg~\cite{Li2022LanguagedrivenSS} to obtain language-aligned image embeddings. We use a voxel size of 2 cm and a batch size of 8 on the ScanNet200.

To pretrain our MEST module, we employ the AdamW optimizer with an initial learning rate of $1\times10^{-4}$, a weight decay of 0.05, and a batch size of 4. Training is conducted for 512 epochs using a polynomial learning-rate scheduler with a decay base of 0.9. We adopt standard data augmentations, including horizontal flipping, random rotations around the z-axis, elastic distortions, and random scaling. We further apply graph-based superpoint clustering~\cite{Landrieu2017LargeScalePC} and use a voxel size of 2~cm on the ScanNet200.

During unified instruction tuning, we apply LoRA~\cite{Hu2021LoRALA} to all linear layers of the LLM backbone, i.e., Vicuna-1.5-7B~\cite{chiang2023vicuna}, except for the final logits head.
We set the LoRA rank to 64 and the scaling factor $\alpha$ to 128.
Optimization is performed using AdamW with a cosine-annealing learning rate schedule, starting from an initial learning rate of $2 \times 10^{-4}$. The batch size is set to 2 per GPU, and model parameters are updated using gradient accumulation with 8 steps.

\section{Additional Ablation Studies}
In this section, we provide additional quantitative analyzes to better understand the behavior of our MEST module. We conduct a series of ablations on three core components of the architecture: the router z-loss weight $\lambda_z$ which regularizes the gating logits, the load-balancing loss weight $\lambda_{blc}$ which encourages more uniform expert utilization and the expert selection strategy (Top-K routing). These experiments offer deeper insights into how different routing regularizers and expert selection mechanisms influence performance across all four benchmarks.

\noindent \textbf{Router z-loss Weight.}
The results in Table~\ref{tab:router_z_loss} show that model performance is relevant to the choice of the router z-loss weight $\lambda_z$.
Without router regularization ($\lambda_z=0$), the model already performs reasonably well. In contrast, applying a moderate amount of router regularization yields clear performance improvements across multiple benchmarks. At $\lambda_z = 1\times10^{-4}$, the model achieves a $+2.4$ mIoU gain on ScanRefer and a $+2.0$ mIoU improvement on Multi3DRefer compared with the baseline. On SQA3D, this setting also produces the strongest EM-R score of $58.9$.
This suggests that the router z-loss effectively prevents extreme routing logits, encouraging more stable and balanced expert utilization.
When the regularization becomes too strong (e.g., $\lambda_z=5\times10^{-4}$), performance begins to degrade, likely due to over-penalizing the gating network, which restricts expert specialization.
Conversely, when $\lambda_z$ is too small ($1\times10^{-5}$ or $1\times10^{-6}$), the regularization is insufficient to improve routing behavior across tasks, resulting in weaker generalization.
Overall, $\lambda_z = 1\times10^{-4}$ offers the best trade-off, delivering the strongest results across four benchmarks.

\begin{table}[t]
\centering
\caption{
Ablation on the router z-loss weight $\lambda_z$ across four benchmarks. Our default setting is highlighted with \colorbox{violet!10}{light violet}.
}
\label{tab:router_z_loss}
\resizebox{\columnwidth}{!}{
\begin{tabular}{c cc cccc cc}
\toprule
\multirow{2}{*}{$\lambda_z$} &
\multicolumn{1}{c}{ScanRefer} &
\multicolumn{1}{c}{Multi3DRefer} &
\multicolumn{4}{c}{ScanQA} &
\multicolumn{2}{c}{SQA3D} \\
\cmidrule(r){2-2}
\cmidrule(r){3-3}
\cmidrule(lr){4-7}
\cmidrule(l){8-9}
 & mIoU↑ & mIoU↑ & B-4↑ & M↑ & R↑ & C↑ & EM↑ & EM-R↑ \\
\midrule
0              & 42.0 & 46.8 & 15.4 & 18.4 & 43.6 & 92.5 & 55.6 & 58.2 \\
$5\times10^{-4}$ & 41.8 & 46.9 & 15.1 & 18.4 & 43.4 & 92.1 & \textbf{56.2} & 58.6 \\
\rowcolor{violet!10}
$1\times10^{-4}$ & \textbf{44.4} & \textbf{48.8} & \textbf{15.8} & 18.4 & 43.5 & 92.7 & 56.0 & \textbf{58.9} \\
$1\times10^{-5}$ & 43.0 & 47.7 & 15.1 & 18.5 & 43.4 & 92.6 & 54.5 & 57.3 \\
$1\times10^{-6}$ & 42.1 & 47.3 & 15.7 & \textbf{18.6} & \textbf{43.7} & \textbf{93.6} & 55.2 & 57.7 \\
\bottomrule
\end{tabular}}
\end{table}

\noindent \textbf{Load-balancing Loss Weight.}
Table~\ref{tab:balancing_loss} shows that the load-balancing loss weight $\lambda_{blc}$ has a relatively weak but consistent regularization effect across all tasks.
Without load-balancing loss (i.e. $\lambda_{blc}=0$), the model already performs competitively.
Introducing a small balancing term ($\lambda_{blc}=1\times10^{-3}$), performance increases by $+1.1$ mIoU and $+1.0$ mIoU on ScanRefer and Multi3DRefer, respectively. For ScanQA, BLEU-4 rises from $15.4$ to $16.0$, accompanied by a slight gain in CIDEr ($92.5$ to $92.8$). A similar trend is observed on SQA3D, where EM-R improves from $58.2$ to $58.4$.
However, compared with the router z-loss $\mathcal{L}_{z}$ (Table~\ref{tab:router_z_loss}), load-balancing loss $\mathcal{L}_{blc}$ has a significantly smaller impact, confirming that it plays an auxiliary rather than decisive role in mixture-of-experts optimization.
When the weight becomes too small ($1 \times 10^{-5}$ or $1 \times 10^{-7}$), the regularization no longer influences the gating network, causing metrics to fall back to or slightly below the baseline.
This indicates that weak balancing is insufficient to meaningfully affect expert utilization, while moderate balancing is beneficial.

\begin{table}[t]
\centering
\caption{
Ablation on the load-balancing loss weight $\lambda_{blc}$ across four benchmarks.
}
\label{tab:balancing_loss}
\resizebox{\columnwidth}{!}{
\begin{tabular}{c cc cccc cc}
\toprule
\multirow{2}{*}{$\lambda_{blc}$} &
\multicolumn{1}{c}{ScanRefer} &
\multicolumn{1}{c}{Multi3DRefer} &
\multicolumn{4}{c}{ScanQA} &
\multicolumn{2}{c}{SQA3D} \\
\cmidrule(r){2-2}
\cmidrule(r){3-3}
\cmidrule(lr){4-7}
\cmidrule(l){8-9}
 & mIoU↑ & mIoU↑ & B-4↑ & M↑ & R↑ & C↑ & EM↑ & EM-R↑ \\
\midrule
0              & 42.0 & 46.8 & 15.4 & 18.4 & \textbf{43.6} & 92.5 & 55.6 & 58.2 \\
$1\times10^{-3}$ & \textbf{43.1} & \textbf{47.8} & \textbf{16.0} & \textbf{18.4} & 43.3 & 92.8 & \textbf{55.9} & \textbf{58.4} \\
$1\times10^{-5}$ & 41.5 & 46.3 & 15.0 & 18.4 & 43.5 & \textbf{93.0} & 55.3 & 57.7 \\
$1\times10^{-7}$ & 42.6 & 47.5 & 15.2 & 18.2 & 43.2 & 92.4 & 55.3 & 57.7 \\
\bottomrule
\end{tabular}}
\end{table}

\noindent \textbf{Joint Ablation of Router z-loss and Load-balancing Loss.}
To better understand the interaction between the router z-loss $\mathcal{L}_{z}$ and the load-balancing loss $\mathcal{L}_{blc}$, we perform a joint ablation in which both regularization terms are varied simultaneously. As shown in Table~\ref{tab:z_and_balance}, combining the two losses does not provide complementary benefits. Using only the router z-loss ($\lambda_z = 1\times10^{-4}$, $\lambda_{blc} = 0$) yields the best overall performance, achieving $44.4$ mIoU on ScanRefer and $48.8$ mIoU on Multi3DRefer. In contrast, applying both losses together slightly harms performance. Specifically, mIoU drops to $42.4$ on ScanRefer and $47.2$ on Multi3DRefer. This indicates that the two regularizers may impose conflicting constraints on the gating network, potentially leading to over-regularization. Using only the load-balancing loss ($\lambda_z=0$, $\lambda_{blc}=1\times10^{-3}$) produces modest improvements over the baseline, but its effect remains notably smaller than that of the router z-loss. Overall, these results show that the router z-loss is the key factor contributing to better training stability by penalizing excessively large gating logits. However, the load-balancing loss offers limited practical benefit. A plausible explanation is that under uneven data distributions, enforcing uniform expert utilization forces experts to update shared parameters across inputs with large domain gaps, introducing gradient interference. This undermines the intended specialization of sparsely-gated MoE layers and ultimately limits performance gains.

\begin{table}[t]
\centering
\caption{
Ablation on router z-loss $\mathcal{L}_{z}$ and load-balancing loss $\mathcal{L}_{blc}$ across four benchmarks.
}
\label{tab:z_and_balance}
\resizebox{\columnwidth}{!}{
\begin{tabular}{cc cc cccc cc}
\toprule
\multicolumn{2}{c}{Losses} &
\multicolumn{1}{c}{ScanRefer} &
\multicolumn{1}{c}{Multi3DRefer} &
\multicolumn{4}{c}{ScanQA} &
\multicolumn{2}{c}{SQA3D} \\
\cmidrule(r){1-2}
\cmidrule(r){3-3}
\cmidrule(r){4-4}
\cmidrule(lr){5-8}
\cmidrule(l){9-10}
$\mathcal{L}_{z}$ & $\mathcal{L}_{blc}$ &
mIoU↑ & mIoU↑ &
B-4↑ & M↑ & R↑ & C↑ &
EM↑ & EM-R↑ \\
\midrule
 &  & 
42.0 & 46.8 & 15.4 & 18.4 & \textbf{43.6} & 92.5 & 55.6 & 58.2 \\
\rowcolor{violet!10}
\checkmark &  &
\textbf{44.4} & \textbf{48.8} & 15.8 & \textbf{18.4} & 43.5 & 92.7 & \textbf{56.0} & \textbf{58.9} \\
 & \checkmark &
43.1 & 47.8 & 16.0 & 18.4 & 43.3 & \textbf{92.8} & 55.9 & 58.4 \\
\checkmark & \checkmark &
42.4 & 47.2 & \textbf{16.1} & 18.3 & 42.9 & 92.3 & 55.0 & 57.6 \\
\bottomrule
\end{tabular}}
\end{table}

\noindent \textbf{Expert Selection Strategy.}
This ablation examines how different expert selection strategies affect model performance, as shown in Table~\ref{tab:init_ablation}. We compare Top-2 routing, equipped with three second-expert activation policies (\textit{All}, \textit{Threshold}, and \textit{Random}), against a deterministic Top-1 routing strategy. Among the Top-2 variants, the \textit{All} policy, which always activates the second-ranked expert, yields the highest grounding performance (43.7/48.1 mIoU on ScanRefer/Multi3DRefer). The \textit{Random} policy, which probabilistically activates the second expert based on its gating score, produces slightly better results on certain ScanQA metrics. The \textit{Threshold} policy, which only enables the second expert when its score exceeds a predefined threshold, is generally more restrictive and tends to underperform relative to the other Top-2 variants. Despite these differences, all three Top-2 strategies consistently fall short of the simpler Top-1 configuration. Even the strongest Top-2 setting lags behind Top-1 across all benchmarks, indicating that activating two experts simultaneously introduces routing redundancy and weakens expert specialization. This reduces the distinctiveness of expert behaviors, ultimately limiting generalization. In contrast, the Top-1 configuration achieves the best overall results, including $44.4$ mIoU on ScanRefer, $48.8$ mIoU on Multi3DRefer, and $56.0$ / $58.9$ EM / EM-R on SQA3D. Routing each token to a single expert leads to clearer specialization, reduced interference during multimodal fusion, and more stable query decoding.

\begin{table}[t]
\centering
\caption{Ablation on the expert selection strategy. We compare different Top-$K$ routing choices (Top-2 vs. Top-1) and several expert initialization strategies for Top-2 routing, including \textit{Threshold}, \textit{Random}, and \textit{All}. Top-1 routing without additional initialization achieves the best performance across all four benchmarks.
}
\label{tab:init_ablation}
\resizebox{\columnwidth}{!}{
\begin{tabular}{l l cc cccc cc}
\toprule
\multirow{2}{*}{Top-$K$} &
\multirow{2}{*}{Method} & 
\multicolumn{1}{c}{ScanRefer} &
\multicolumn{1}{c}{Multi3DRefer} &
\multicolumn{4}{c}{ScanQA} &
\multicolumn{2}{c}{SQA3D} \\
\cmidrule(r){3-3}
\cmidrule(r){4-4}
\cmidrule(lr){5-8}
\cmidrule(l){9-10}
 &  & mIoU↑ & mIoU↑ & B-4↑ & M↑ & R↑ & C↑ & EM↑ & EM-R↑ \\
\midrule
\multicolumn{1}{c|}{\multirow{3}{*}{Top-2}} 
 & Threshold & 41.1 & 45.9 & 15.1 & 18.6 & 43.7 & 93.4 & 55.9 & 58.6 \\
\multicolumn{1}{c|}{} 
 & Random    & 42.7 & 47.7 & \textbf{16.4} & \textbf{18.8} & \textbf{43.8} & \textbf{94.3} & 55.4 & 57.9 \\
\multicolumn{1}{c|}{} 
 & All       & 43.7 & 48.1 & 15.9 & 18.3 & 43.1 & 92.3 & 55.0 & 57.7 \\
\midrule
\rowcolor{violet!10}
\multicolumn{1}{c|}{Top-1} 
 & None      & \textbf{44.4} & \textbf{48.8} & 15.8 & 18.4 & 43.5 & 92.7 & \textbf{56.0} & \textbf{58.9} \\
\bottomrule
\end{tabular}}
\end{table}

\section{Additional Qualitative Results}

\begin{figure*}[t]
    \centering
    \includegraphics[width=0.85\textwidth]{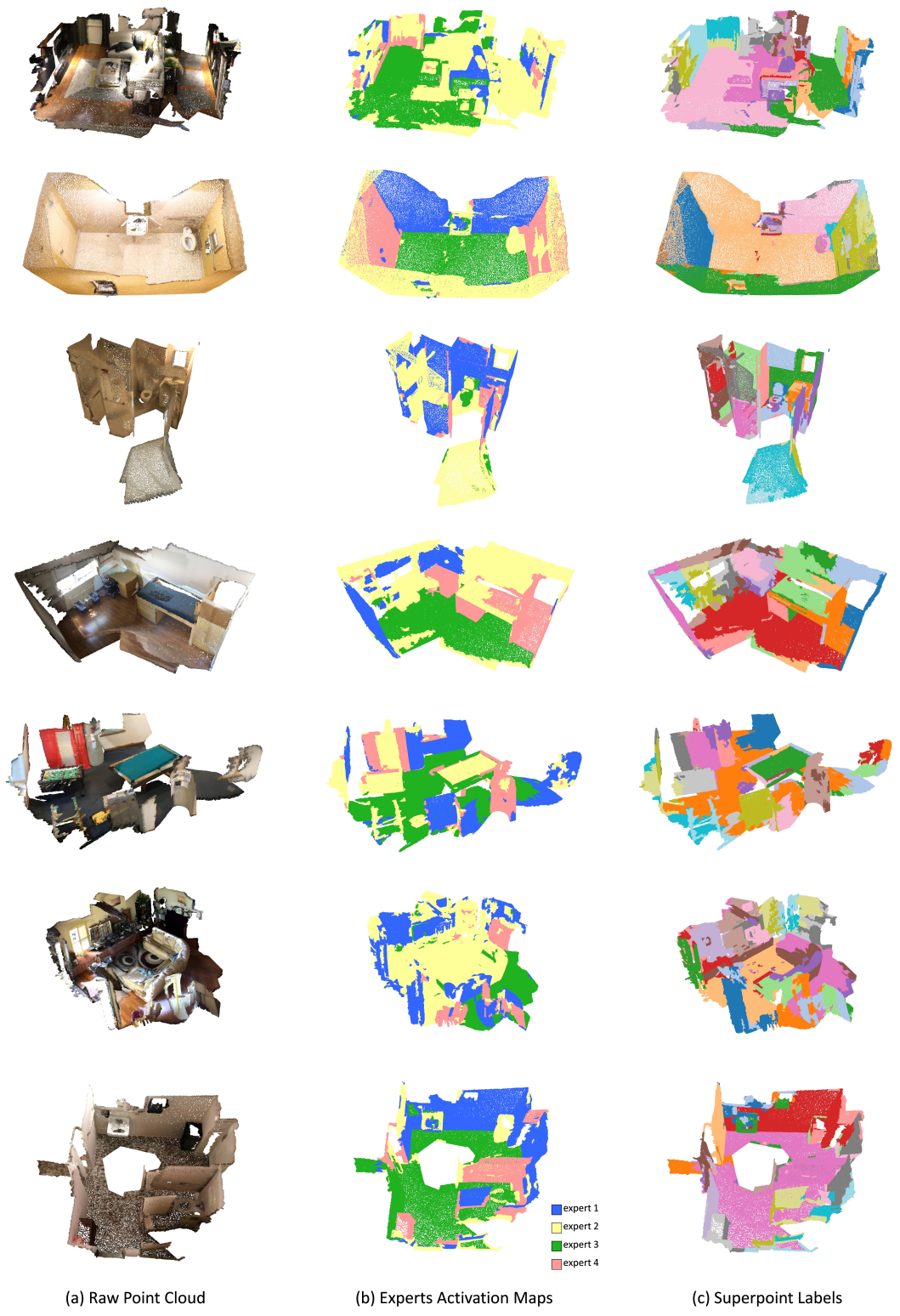}
    \caption{Qualitative visualization of expert specialization. 
    (a) Raw point cloud.
    (b) Experts activation maps produced by our \algorithmname~, where each color corresponds to the dominant expert assigned to each point. 
    (c) Superpoint labels used for training. } 
    \label{fig:experts_activation}
    \vspace{-3mm}
\end{figure*}

Fig.~\ref{fig:experts_activation} provides qualitative visualizations that illustrate how our \algorithmname~interacts with the multi-modalities of 3D scenes.
It is evident that different experts specialize in distinct modality of the scene in the expert activation maps.

\begin{figure*}[t]
    \centering
    \includegraphics[width=0.85\textwidth]{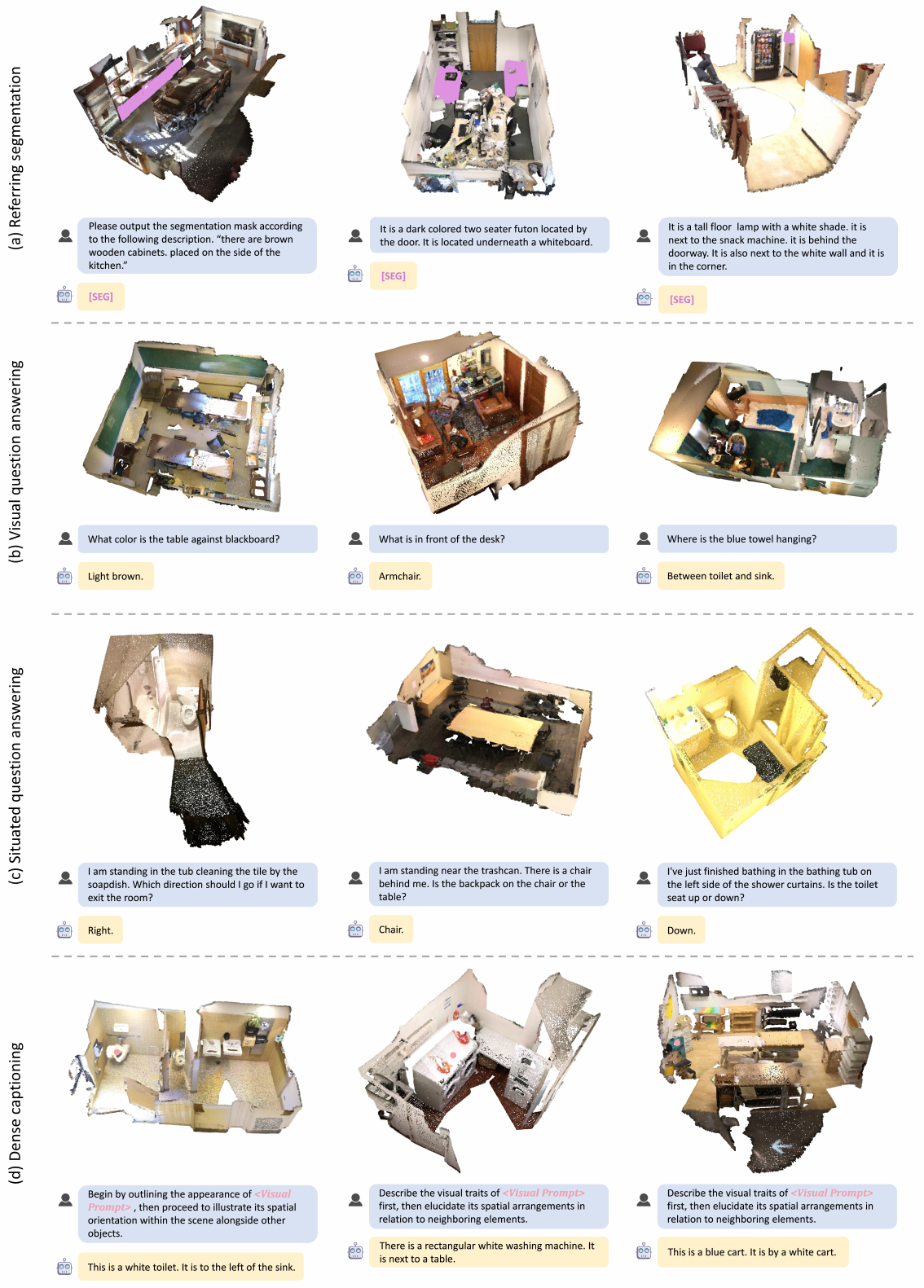}
    \caption{
        Qualitative results across four 3D scene understanding tasks. 
        (a) Referring segmentation: examples from kitchen, office, and lounge scenes, where \algorithmname~accurately segments the target objects described in natural language. 
        (b) Visual question answering: questions involving object color, category, and spatial relations, with answers grounded in the 3D geometry. 
        (c) Situated question answering: queries that require reasoning about direction, location, or object status from an embodied perspective. 
        (d) Dense captioning: descriptions generated for different types of objects across different rooms. 
    }

    \label{fig:multitask_vis}
    \vspace{-3mm}
\end{figure*}

Fig.~\ref{fig:multitask_vis} presents qualitative results across four 3D scene understanding tasks, including referring segmentation, visual question answering, situated question answering, and dense captioning, demonstrating the unified capability of our \algorithmname~ framework to handle diverse multimodal 3D scene understanding tasks within a single model.

\end{document}